# Corn leaf detection using Region based convolutional neural network

○Mohammad Ibrahim sarker [1], , Heechan Yang [2], Hyongsuk Kim [3]
[1] 전북대학교 전자공학부 (TEL: 063-270-2477; E-mail: sarkeribrahim@gmail.com)
[2] 전북대학교 전자공학부 (TEL: 063-270-2477; E-mail: yhc3006@naver.com)
[3] 전북대학교 전자공학부, 지능형로봇연구센터 (TEL: 063-270-2477; E-mail: hskim@jbnu.ac.kr)

**Abstract** The field of machine learning has become an increasingly budding area of research as more efficient methods are needed in the quest to handle more complex image detection challenges. To solve the problems of agriculture is more and more important because food is the fundamental of life. However, the detection accuracy in recent corn field systems are still far away from the demands in practice due to a number of different weeds. This paper presents a model to handle the problem of corn leaf detection in given digital images collected from farm field. Based on results of experiments conducted with several state-of-the-art models adopted by CNN, a region-based method has been proposed as a faster and more accurate method of corn leaf detection. Being motivated with such unique attributes of ResNet, we combine it with region based network (such as faster rcnn), which is able to automatically detect corn leaf in heavy weeds occlusion. The method is evaluated on the dataset from farm and we make an annotation ourselves. Our proposed method achieves significantly outperform in corn detection system.

## 1. Introduction

As world population continues to grow, land and natural resources have continued to diminish as well. As a result, precision agriculture has now become the focus for most researchers. In the agricultural sector, the control of weeds is not just a costly activity but it also consumes a lot of time. In addition, the use of herbicides over an extended period of time causes pollution and consequently poses a hazard to humans, animals as well as the environment. It is a known fact that the usage of herbicides in farms has been going on consistently for a very long time now and has contributed to environmental pollution.

Therefore, in a bid to save herbicides and reduce environmental pollution without reducing crop yield, research efforts are now geared towards developing technologies in which selected herbicides can be used to target specific weeds. The use of artificial intelligence in the application of herbicides to target weeds in a precise and efficient manner is one attractive alternative. However, it is still difficult for machines to locate and identify more than one item at a time.

However, research in the area of object detection has yielded significant positive results with the use of convolutional neuronal networks (CNNs) in the last couple of years. There has been debate about the pros and cons between shallow networks and deep neural networks in machine learning and it has been seen that deeper networks do not need as much components as shallow networks.

In recent times, object detection networks consist of two parts. The first part is the classification of the object after which a boundary box is made to make the object classified. Convolutional network with MLP and boundary box regression have been employed by researchers to classify the object. A lot of other models such as R-CNN, Fast R-CNN, Faster R-CNN, and R-FCN can also be used.

In our proposed approach, we focus on the detection of corn leaf in heavy weeds occlusion. The proposed system

※ This work was carried out with the support of " Development of real-time diagnosis analysis technology for the major disease and insects of tomato using neural networks (Project No. PJ0120642017)", NRF (project No.-2016R1A2B4015514 BK 21 Plus and Intelligent Robot Research Center, Chonbuk National University, Republic of Korea.

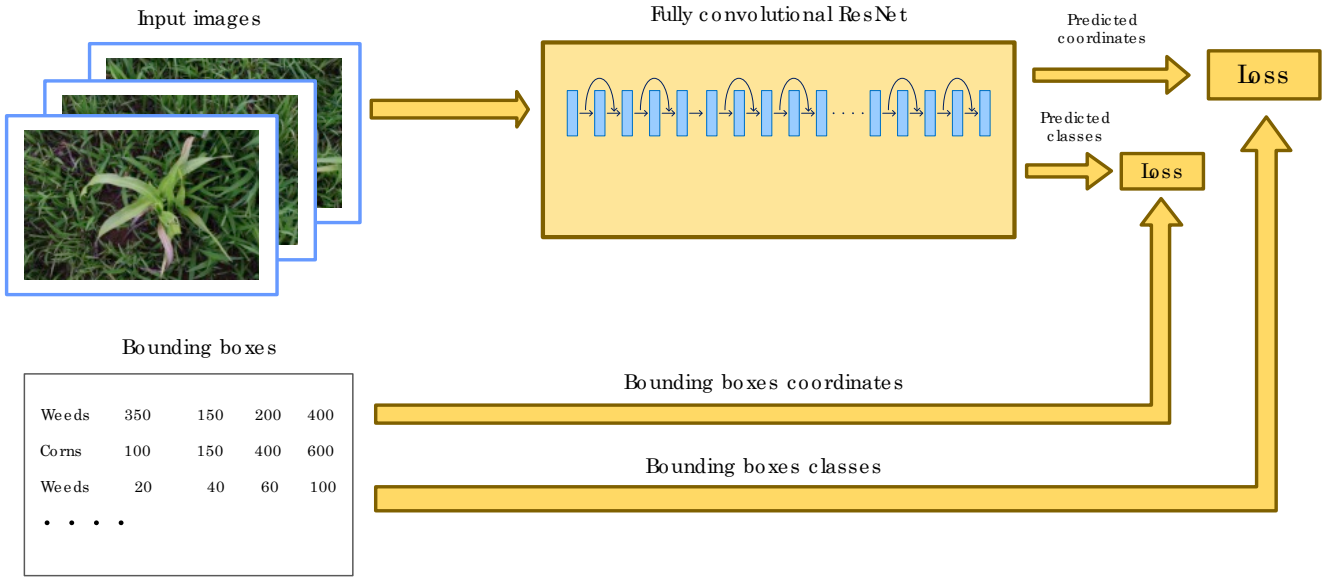

*Figure 1: Proposed network*

introduces a practical and applicable solution for detection corn leaf and weeds which in fact represents a main comparable difference with traditional methods for corn leaf and weeds classification.

## 2. Materials and Methods

In order to train the networks to be able to detect corn leaf as well as weeds, a vast amount of training images are required. While doing annotation first we need to mark corn leaf and weed instance in the image and denote the names manually. The images are subsequently scaled down to 1200*1100 pixels. This size limit is because the Nvidia Titan X GPU used for training would otherwise run low on memory. We do data augmentation for increasing our dataset as well as improving performance by doing randomly croping, Rotate left and right, zoom, Stretch vertically & horizontally, Zoom in / out, Elastic deformations.

*2.1 Network architecture*

For running a detection and recognition pipeline, we usually faced with a challenge of vast number of different positions and scales in which object we are interested might be in an image. Since passing an image through recognition network is quite time-consuming due to many computations in convolutional layers, many previous approaches tried to overcome this challenge with splitting this task into two different ones – detection of Regions of Interest (RoI) and then classifying only a limited number of Roi's by convolutional networks. However since many computations between object detection and recognition can be shared, Faster R-CNN architecture merges those two computations back into one using a concept of Region Proposal Network which is a small extension of original recognition network. The RPN is put right before last pooling layer and consists of convolutional layer of size $3 \times 3$ (where width dimension is equal to width of preceding layer) and feeds the output of this layer into two sibling layers–box-regression layer (reg) and box-classification layer(cls) which are both implemented as a convolutional layers of dimension $1 \times 1$. The reg and cls layers outputs 4k and 2k dimensional vectors respectively, where k stands for number of so called anchors at the processed position. Each anchor stands for a different setting of scales and aspect ratios at which RPN is analyzing given position. The original paper used three different scales and three different aspect ratios resulting in k = 9. This approach ensures translational invariation of anchors. Reg layer estimates regression of possible bounding box of an object and class layer computes 2 probabilities of object/not-object in each anchor at given position.

Outputs of this RPN network are then fed into Fast R-CNN part of networks, which expects RoI proposals and feature maps of last convolutional layer and applies so called RoI pooling effectively replacing last max pooling layer. RoI pooling divides each region of RoI proposal of dimension $w \times h$ into grid of size $W \times H$ where W and H are parameters of a layer (and in our experiments W = H = 7 was used) and performing classical max pooling on each cell of size $w/W \times h/H$ effectively squashing features of any RoI into fixed length vector of size $W \times H$. This RoI pooled vector is then fed into usual fully-connected layers where the output at the end actually consists of two different output layers, one computing probability of a given class similarly as last layer of Resnet network and the other one estimating positions of bounding box depending on found class, therefore having 4C neurons where C is number of classes to classify. Typically our proposed network can be trained in an end-to-end approach. Fig. 1. Shows the overview system of corn leaf and weeds detection.

## 3. Results

To evaluate the performance on the corn leaf dataset, we compute the mean average precision (mAP). AP is the area under the Precision-Recall curve, which is a very popular performance measure in information retrieval. The corn leaf detection is considered true or false according to its overlap with the ground-truth bounding box. If the overlap score is more than 0.5, it will be considered as positive, otherwise negative. The overlap score between two boxes is defined as:

$$IoU = \frac{GT \cap DET}{GT \cup DET} \quad (1)$$

Where the axis is aligned bounding rectangle around area ground-truth bounding box and is the detected bounding box. Our proposed method is evaluated on a 64 bits Ubuntu 11.04 computer with 4 Titan X GPUs. Table 1 shows that in our task Resnet 101 performs better than VGG16.

Table 1. Comparisons on corn leaf and weeds dataset using Different Model.

| Architecture | Dataset | mAP |
|---|---|---|
| Faster R-CNN (VGG 16) | Corn leaf and weeds | 0.6474 |
| Faster R-CNN (ResNet-101) | Corn leaf and weeds | 0.7036 |

*3.1 Qualitative visualization*

All the dataset come from same farm. To make test dataset different, we do zoom in the images first, then we do a test. Because almost all dataset are taken from same angle when take photo of them. After doing a zoom, the dataset will become more difficult to do detection since some image become occlusion as shown in figure 2, second and third columns.

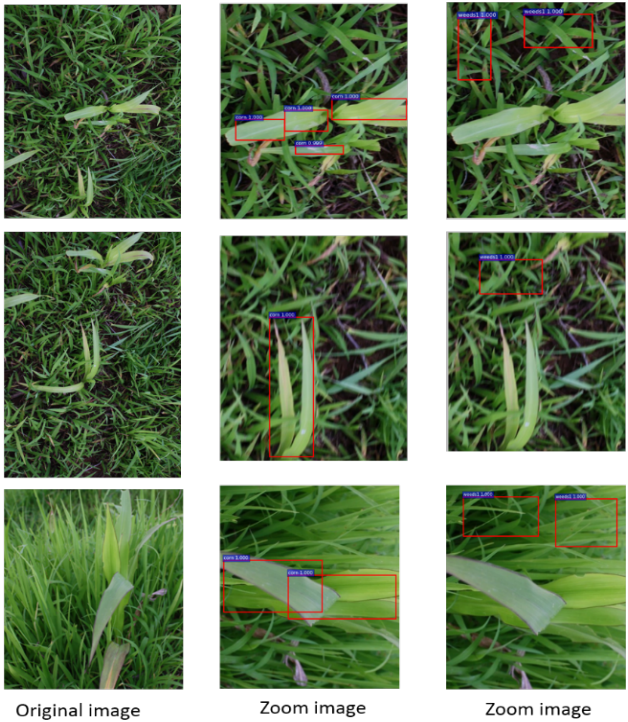

**Figure 2:** Detection results of corn leaf and weeds. First column is the original image, second and last column are zoom images.

## 4. Conclusion:

In this paper a method capable of detecting corn leaf in highly occluded by weeds has been presented. Current methods for corn leaf had trouble with detecting weeds instances in cereal fields due to heavy weeds occlusion. We addressed this problem of detecting corn leaf in cereal fields by using a region based convolutional neural network that is based on a fine-tuned version of the Faster R-CNN architecture. The network has been trained on only limited number of annotated images.
Results show that the algorithm is able to detect 70% of the weeds in the corn field, despite the fact that large parts of the weeds overlap with corn plants. The algorithm, however, encounters problems in detecting very small weeds, grasses, and weeds exposed to a severe degree of overlap.